\documentclass[letterpaper, 10 pt, conference]{ieeeconf}  

\IEEEoverridecommandlockouts
\usepackage{url}
\usepackage{balance}
\usepackage{amsmath}
\usepackage{amssymb,bm}
\usepackage{amsfonts}
\usepackage{fancyvrb}
\usepackage{enumerate}
\usepackage{tabulary}
\usepackage{multirow}
\usepackage{cite}
\usepackage{lipsum}  
\usepackage{graphicx}
\usepackage{epstopdf}
\usepackage[misc]{ifsym}
\usepackage{color}
\usepackage{xcolor}
\usepackage{xxcolor}
\usepackage{pgf}
\usepackage{pgfplots}
\usepackage{tikz}
\usepackage{float}
\usepackage{lipsum}
\usepackage{booktabs}
\usepackage{verbatim}
\usepackage{hyperref}
\usepackage{makecell}
\usepackage{xfrac}
\usepackage{siunitx}
\usepackage{colortbl}
\usepackage{forest}

\usetikzlibrary{external}
\usepackage{pifont}

\usepackage{subcaption}

\usepgfplotslibrary{groupplots,fillbetween}
\usepackage{pgfplotstable}

\usepackage{ifthen}
\usepackage{forloop}
\usepackage{listings}
\usepackage[lined,boxed,commentsnumbered,linesnumbered]{algorithm2e}

\definecolor{codegreen}{rgb}{0,0.6,0}
\definecolor{codepurple}{rgb}{0.58,0,0.82}
\definecolor{backcolour}{rgb}{0.95,0.95,0.92}
\lstdefinestyle{buzz}{
    backgroundcolor=\color{black!5},   
    commentstyle=\color{codegreen},
    keywordstyle=\color{blue},
    numberstyle=\tiny\color{black!30},
    stringstyle=\color{codepurple},
    basicstyle=\footnotesize\ttfamily,
    breakatwhitespace=false,         
    breaklines=true,                 
    captionpos=b,                    
    keepspaces=true,                 
    numbers=left,                    
    numbersep=5pt,                  
    showspaces=false,                
    showstringspaces=false,
    showtabs=false,                  
    tabsize=2,
}
\SetAlCapSkip{0.5em}

\title{\LARGE \bf
Race Against the Machine: a Fully-annotated, Open-design\\ Dataset of Autonomous and Piloted High-speed Flight
}

\author{
Michael Bosello,
Davide Aguiari,
Yvo Keuter,
Enrico Pallotta,
Sara Kiade,
Gyordan Caminati,\\
Flavio Pinzarrone,
Junaid Halepota,
Jacopo Panerati,
and Giovanni Pau
\thanks{All the authors are with the Autonomous Robotics Research Center of the Technology Innovation Institute, Abu Dhabi, United Arab Emirates.
Michael Bosello and Giovanni Pau also are with the University of Bologna, Bologna, Italy.
E-mails:
        {\tt \{firstname.lastname\}@tii.ae}}}

\begin{document}
\maketitle
\thispagestyle{empty}
\pagestyle{empty}

\begin{abstract}
Unmanned aerial vehicles, and multi-rotors in particular, can now perform dexterous tasks in impervious environments, from infrastructure monitoring to emergency deliveries.
Autonomous drone racing has emerged as an ideal benchmark to develop and evaluate these capabilities.
Its challenges include accurate and robust visual-inertial odometry during aggressive maneuvers, complex aerodynamics, and constrained computational resources.
As researchers increasingly channel their efforts into it, they also need the tools to timely and equitably compare their results and advances.
With this dataset, we want to \emph{(i)} support the development of new methods and \emph{(ii)} establish quantitative comparisons for approaches originating from the broader robotics and artificial intelligence communities.
We want to provide a one-stop resource that is comprehensive of \emph{(i)} aggressive autonomous and piloted flight, \emph{(ii)} high-resolution, high-frequency visual, inertial, and motion capture data, \emph{(iii)} commands and control inputs, \emph{(iv)} multiple light settings, and \emph{(v)} corner-level labeling of drone racing gates.
We also release the complete specifications to recreate our flight platform, using commercial off-the-shelf components and the open-source flight controller Betaflight, to democratize drone racing research.
Our dataset, open-source scripts, and drone design are available at: \href{https://github.com/tii-racing/drone-racing-dataset}{\texttt{github.com/tii-racing/drone-racing-dataset}}.
\end{abstract}

\section{Introduction}
\label{sec:introduction}

Unmanned aerial vehicles (UAVs) and multi-rotor drones have become ubiquitous robotic platforms, supporting a wide array of industries from video-making to warehouse monitoring, to surveillance and inspection of energy and transport infrastructure. Drone racing, in particular, has emerged as the go-to benchmark problem to measure {the advances made by researchers in the quest to surpass human-level, autonomous performance in fast and aggressive flight~\cite{scaramuzza-nature,romero2022model,delmerico2019,scaramuzza-scirob}}.
Yet, drone racing competitions, equipment, and venues can still be difficult and expensive to access.  

The last decade of machine learning progress has shown how datasets, open standards, and open-source code help scientific progress and transparency, shaping the entire scientific fields~\cite{imagenet}. One of the fundamental advantages of datasets is to greatly simplify and shorten the development pipeline of new methods by allowing researchers to re-use the tried and tested data collection and consolidation work of others.
Datasets allow researchers from different parts of the world and disciplines to work on common problems.

\begin{figure}
    \includegraphics[trim={0cm 10cm 0cm 10cm},clip,width=0.48\textwidth]{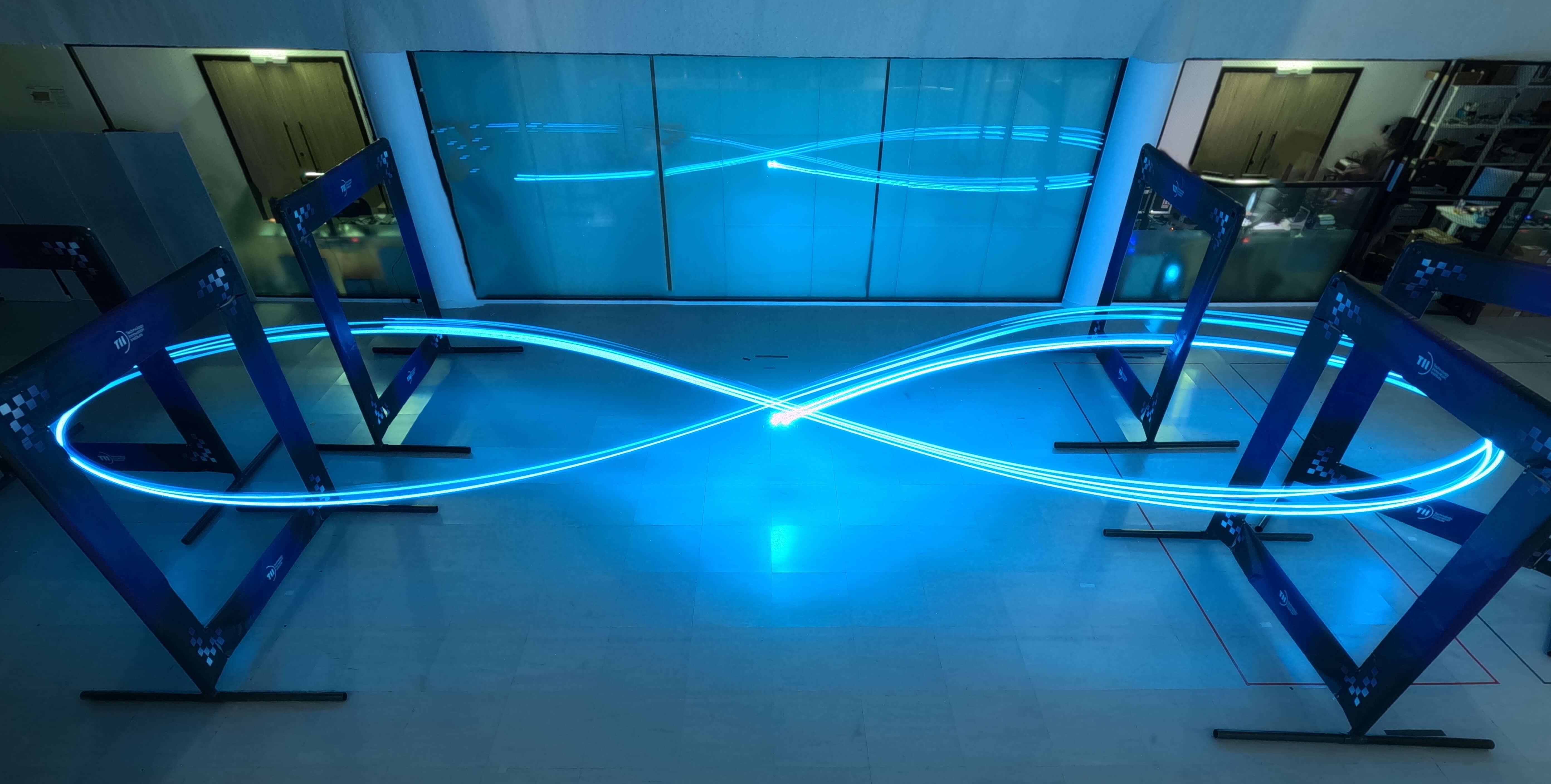}
    \includegraphics[trim={0cm 12cm 0cm 28cm},clip,width=0.48\textwidth,]{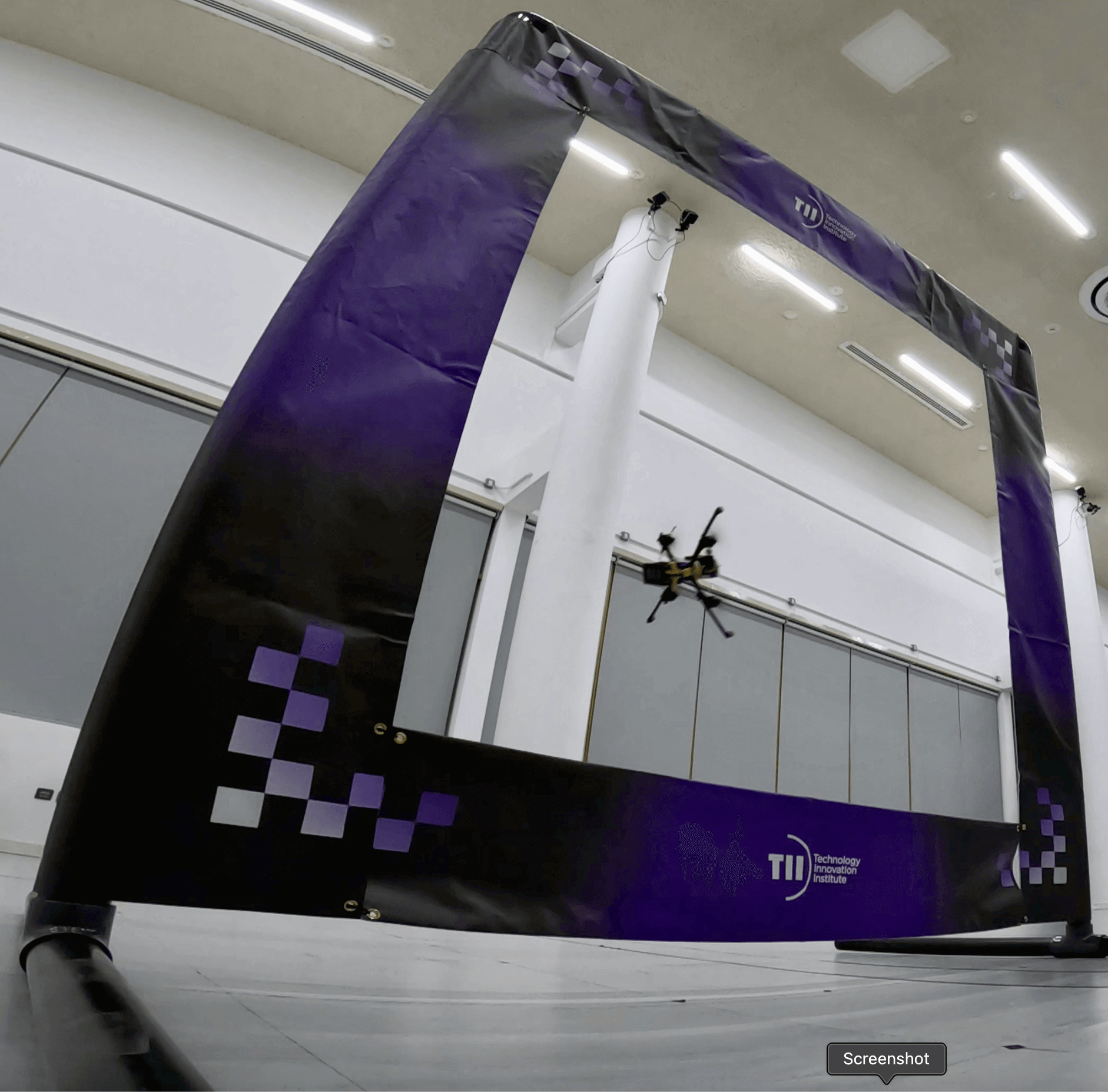}
    \vspace{-.5em}
    \caption{
    Long exposure, low-light capture of the open-design racing drone used to collect the dataset (top), and an aggressive maneuver through one of the labeled gates (bottom).
    }
    \label{fig:beautypicA}
\end{figure}

\begin{table*}[h!]
    \centering
    \caption{
    Comparison of multi-rotor and drone racing datasets for visual-inertial odometry, scene understanding, and control
    }
    \newcommand{\cmark}{\ding{51}}\newcommand{\xmark}{\ding{55}}

\centering
\scalebox{0.725}{\hspace{-0.0cm}\begin{tabular}{
>{\centering}p{.9cm}
>{\centering}p{.95cm}
>{\centering}p{.75cm}
>{\centering}p{.9cm}
>{\centering}p{.9cm}
>{\centering}p{.5cm}
>{\centering}p{.6cm} 
>{\centering}p{.9cm} >{\centering}p{1.9cm}
>{\centering}p{.9cm}	
>{\centering}p{.6cm} 
>{\centering}p{.5cm}
>{\centering}p{.5cm}
>{\centering}p{1cm} 
>{\centering}p{1cm}
>{\centering}p{1cm}
>{\centering}p{1cm}
>{\centering}p{1cm}
>{\centering}p{1.3cm} 
}
    \toprule

\multirow{2}{*}{{\scriptsize Ref.}}
    & {\scriptsize Time \&}
    & {\scriptsize Data}
    & \multicolumn{2}{c}{\scriptsize Conditions}
    & \multicolumn{2}{c}{\scriptsize Gates}
    & {\scriptsize Top}
    & \multicolumn{5}{c}{\scriptsize Vision/Camera Specifications}
    & \multicolumn{2}{c}{\scriptsize Pose/Inertial Data}
    & \multicolumn{2}{c}{\scriptsize Control Inputs}
    & {\scriptsize Battery}
    & {\scriptsize Data}
    \tabularnewline
    
& {\scriptsize Distance}
    & {\scriptsize Coll.}
    & {\scriptsize Scene}
    & {\scriptsize Lighting}
    & {\scriptsize Pose}
    & {\scriptsize Labels}
    & {\scriptsize Speed}
    & {\scriptsize Resolution/Freq.}
    & {\scriptsize Color}
    & {\scriptsize FoV}
& {\scriptsize Stereo}
    & {\scriptsize \textcolor{black}{Event}}
    & {\scriptsize IMU}
    & {\scriptsize MoCap}
    & {\scriptsize CTBR}
    & {\scriptsize Motor}
    & {\scriptsize Voltage}
    & {\scriptsize Formats}
    \tabularnewline
    
    \cmidrule(lr){1-1}
    \cmidrule(lr){2-2}
    \cmidrule(lr){3-3}
    \cmidrule(lr){4-5}
    \cmidrule(lr){6-7}
    \cmidrule(lr){8-8}
    \cmidrule(lr){9-13}
    \cmidrule(lr){14-15}
    \cmidrule(lr){16-17}
    \cmidrule(lr){18-18}
    \cmidrule(lr){19-19}

    \cellcolor{black!10} {\scriptsize Ours \tiny{{TII-RATM}}}
& \cellcolor{black!0} {\scriptsize \textcolor{black}{$\sim$29'} $\sim$7km} & \cellcolor{green!10} {\scriptsize Real} & \cellcolor{black!0} {\scriptsize Indoor} & \cellcolor{green!10} {\scriptsize 3 Levels Labeled} & \cellcolor{green!10} {\scriptsize \cmark} & \cellcolor{green!10} {\scriptsize \cmark\hyperlink{hyref:01}{$^\dag$}} & \cellcolor{green!10} {\scriptsize  9.5m/s\hyperlink{hyref:99}{$^{\diamond}$} 21.8m/s\hyperlink{hyref:09}{$^{\P}$}} & \cellcolor{green!10} {\scriptsize 640x480@120Hz} & \cellcolor{green!10} {\scriptsize RGB} & \cellcolor{black!0} {\scriptsize \textcolor{black}{D~}175$^\circ$} & \cellcolor{red!10} {\scriptsize \xmark} & \cellcolor{red!10} {\scriptsize \xmark} & \cellcolor{green!10} {\scriptsize @500Hz} & \cellcolor{green!10} {\scriptsize @275Hz} & \cellcolor{green!10} {\scriptsize @100Hz} & \cellcolor{green!10} {\scriptsize @100Hz} & \cellcolor{green!10} {\scriptsize @50Hz} & \cellcolor{black!0} {\scriptsize \href{https://github.com/tii-racing/drone-racing-dataset/releases/tag/v2.0.0}{\texttt{rosbag}, CSV, \textcolor{black}{JPEG}}} \tabularnewline
    
    \cmidrule(lr){2-19}

    {\scriptsize \cite{delmerico2019} \tiny{{UZH-FPV}}} & \cellcolor{black!0} {\scriptsize \textcolor{black}{$\sim$24' $\sim$11km}} & \cellcolor{green!10} {\scriptsize Real} & \cellcolor{black!0} {\scriptsize Indoor; Outdoor} & \cellcolor{black!10} {\scriptsize \textcolor{black}{Multiple, Unlabel.}} & \cellcolor{red!10} {\scriptsize \xmark} & \cellcolor{red!10} {\scriptsize \xmark} & \cellcolor{green!10} {\scriptsize \textcolor{black}{26.8m/s}\hyperlink{hyref:09}{$^{\P}$} 23.4m/s\hyperlink{hyref:99}{$^{\diamond}$}} & \cellcolor{black!10} {\scriptsize \textcolor{black}{848x800@30Hz} 346x260@50Hz 640x480@\textcolor{black}{30Hz}\hyperlink{hyref:02}{$^*$}} & \cellcolor{black!10} {\scriptsize \textcolor{black}{Grayscale} Grayscale Grayscale} & \cellcolor{black!0} {\scriptsize \textcolor{black}{D~163$^\circ$} 120$^\circ$ 186$^\circ$} & \cellcolor{green!10} {\scriptsize \textcolor{black}{\cmark}} & \cellcolor{green!10} {\scriptsize \cmark} & \cellcolor{green!10} {\scriptsize \textcolor{black}{@200Hz} @500Hz @1000Hz} & \cellcolor{black!10} {\scriptsize \textcolor{black} {\hspace{.5cm} @20Hz\hyperlink{hyref:07}{$^{\S}$}}} & \cellcolor{red!10} {\scriptsize \xmark} & \cellcolor{red!10} {\scriptsize \xmark} & \cellcolor{red!10} {\scriptsize \xmark} & \cellcolor{black!0} {\scriptsize \href{https://fpv.ifi.uzh.ch/datasets/}{\texttt{rosbag}, TXT, \textcolor{black}{PNG}}} \tabularnewline
    
    \cmidrule(lr){2-19}

    {\scriptsize \textcolor{black}{\cite{alphapilotchallenge}} \tiny{{AlphaPilot}}} & \cellcolor{black!0} {\scriptsize \textcolor{black}{n/a\hyperlink{hyref:06}{$^\ddag$}}} & \cellcolor{green!10} {\scriptsize Real} & \cellcolor{black!0} {\scriptsize Indoor, 1 Gate} & \cellcolor{black!10} {\scriptsize \textcolor{black}{Multiple, Unlabel.}} & \cellcolor{red!10} {\scriptsize \xmark} & \cellcolor{green!10} {\scriptsize \cmark\hyperlink{hyref:02}{$^{||}$}} & \cellcolor{red!10} {\scriptsize \textcolor{black}{n/a}} & \cellcolor{black!10} {\scriptsize 1296x864} & \cellcolor{green!10} {\scriptsize RGB} & \cellcolor{black!0} {\scriptsize \textcolor{black}{n/a}} & \cellcolor{red!10} {\scriptsize \xmark} & \cellcolor{red!10} {\scriptsize \xmark} & \cellcolor{red!10} {\scriptsize \xmark} & \cellcolor{red!10} {\scriptsize \xmark} & \cellcolor{red!10} {\scriptsize \xmark} & \cellcolor{red!10} {\scriptsize \xmark} & \cellcolor{red!10} {\scriptsize \xmark} & \cellcolor{black!0} {\href{https://www.herox.com/alphapilot/85-2019-virtual-qualifier-tests}{\scriptsize JSON, \textcolor{black}{JPEG}}} \tabularnewline
    
    \cmidrule(lr){2-19}

    {\scriptsize \cite{antonini2020} \tiny{{Blackbird}}} & \cellcolor{black!0} {\scriptsize $\sim$10h \textcolor{black}{$\sim$100km}} & \cellcolor{black!10} {\scriptsize \textcolor{black}{Real + Synth.}} & \cellcolor{black!0} {\scriptsize Indoor, 5 Scenes} & \cellcolor{black!10} {\scriptsize \textcolor{black}{Multiple, Unlabel.}} & \cellcolor{red!10} {\scriptsize \xmark} & \cellcolor{red!10} {\scriptsize \xmark} & \cellcolor{black!10} {\scriptsize 7m/s} & \cellcolor{green!10} {\scriptsize 1024x768@120Hz\hyperlink{hyref:04}{$^{\dag\dag}$} 1024x768@360Hz\hyperlink{hyref:04}{$^{\dag\dag}$}} & \cellcolor{green!10} {\scriptsize Grayscale RGB} & \cellcolor{black!0} {\scriptsize \textcolor{black}{V~}60$^\circ$} & \cellcolor{green!10} {\scriptsize \cmark} & \cellcolor{red!10} {\scriptsize \xmark} & \cellcolor{green!10} {\scriptsize @100Hz} & \cellcolor{green!10} {\scriptsize @360Hz} & \cellcolor{red!10} {\scriptsize \xmark} & \cellcolor{green!10} {\scriptsize @190Hz} & \cellcolor{red!10} {\scriptsize \xmark} & \cellcolor{black!0} {\scriptsize \href{https://github.com/mit-aera/Blackbird-Dataset}{\texttt{rosbag}, CSV, \textcolor{black}{MP4}, PNG Depth}} \tabularnewline
    
    \cmidrule(lr){2-19}

    {\scriptsize \cite{burri2016} \tiny{{EuRoC}}} & \cellcolor{black!0} {\scriptsize $\sim$22' $\sim$1km} & \cellcolor{green!10} {\scriptsize Real} & \cellcolor{black!0} {\scriptsize Indoor, 2 Scenes} & \cellcolor{black!10} {\scriptsize \textcolor{black}{Multiple, Unlabel.}} & \cellcolor{red!10} {\scriptsize \xmark} & \cellcolor{red!10} {\scriptsize \xmark} & \cellcolor{red!10} {\scriptsize 2.3m/s} & \cellcolor{red!10} {\scriptsize 752x480@20Hz} & \cellcolor{black!10} {\scriptsize Grayscale} & \cellcolor{black!0} {\scriptsize \textcolor{black}{H~}115$^\circ$} & \cellcolor{green!10} {\scriptsize \cmark} & \cellcolor{red!10} {\scriptsize \xmark} & \cellcolor{green!10} {\scriptsize @200Hz} & \cellcolor{green!10} {\scriptsize @20Hz\hyperlink{hyref:07}{$^{\S}$} @100Hz} & \cellcolor{red!10} {\scriptsize \xmark} & \cellcolor{red!10} {\scriptsize \xmark} & \cellcolor{red!10} {\scriptsize \xmark} & \cellcolor{black!0} {\scriptsize \href{https://projects.asl.ethz.ch/datasets/doku.php?id=kmavvisualinertialdatasets}{CSV, PLY, PNG}} \tabularnewline
    
    \cmidrule(lr){2-19}

    {\scriptsize \cite{upenn2018} \tiny{{GRASP}}} & \cellcolor{black!0} {\scriptsize \textcolor{black}{$\sim$10' $\sim$3km}} & \cellcolor{green!10} {\scriptsize \textcolor{black}{Real}} & \cellcolor{black!0} {\scriptsize \textcolor{black}{Outdoor; 1 Scene}} & \cellcolor{black!10} {\scriptsize \textcolor{black}{Multiple, Unlabel.}} & \cellcolor{red!10} {\scriptsize \textcolor{black}{\xmark}} & \cellcolor{red!10} {\scriptsize \textcolor{black}{\xmark}} & \cellcolor{green!10} {\scriptsize \textcolor{black}{17.5m/s}} & \cellcolor{black!10} {\scriptsize \textcolor{black}{960x800}@40Hz} & \cellcolor{black!10} {\scriptsize \textcolor{black}{Grayscale}} & \cellcolor{black!0} {\scriptsize \textcolor{black}{n/a}} & \cellcolor{green!10} {\scriptsize \textcolor{black}{\cmark}} & \cellcolor{red!10} {\scriptsize \textcolor{black}{\xmark}} & \cellcolor{green!10} {\scriptsize \textcolor{black}{@200Hz}} & \cellcolor{red!10} {\scriptsize \textcolor{black}{\xmark}} & \cellcolor{red!10} {\scriptsize \textcolor{black}{\xmark}} & \cellcolor{red!10} {\scriptsize \textcolor{black}{\xmark}} & \cellcolor{red!10} {\scriptsize \textcolor{black}{\xmark}} & \cellcolor{black!0} {\scriptsize \href{https://github.com/KumarRobotics/msckf_vio/wiki/Dataset}{\texttt{rosbag}}} \tabularnewline
    
    \cmidrule(lr){2-19}

    {\scriptsize \cite{pfeiffer2021} \tiny{{EyeGaze}}} & \cellcolor{black!0} {\scriptsize \textcolor{black}{$\sim$300' $\sim$100}km} & \cellcolor{red!10} {\scriptsize \textcolor{black}{Synth.}} & \cellcolor{black!0} {\scriptsize Indoor, 2 Scenes} & \cellcolor{black!10} {\scriptsize \textcolor{black}{Multiple, Unlabel.}} & \cellcolor{green!10} {\scriptsize \cmark} & \cellcolor{green!10} {\scriptsize \cmark\hyperlink{hyref:05}{$^{\P\P}$}} & \cellcolor{green!10} {\scriptsize 13.8m/s} & \cellcolor{black!10} {\scriptsize 800x600@60Hz} & \cellcolor{green!10} {\scriptsize RGB} & \cellcolor{black!0} {\scriptsize 120$^\circ$} & \cellcolor{red!10} {\scriptsize \xmark} & \cellcolor{red!10} {\scriptsize \xmark} & \cellcolor{red!10} {\scriptsize \xmark} & \cellcolor{green!10} {\scriptsize @500Hz\hyperlink{hyref:08}{$^{**}$}} & \cellcolor{green!10} {\scriptsize \textcolor{black}{@500Hz}} & \cellcolor{red!10} {\scriptsize \xmark} & \cellcolor{red!10} {\scriptsize \xmark} & \cellcolor{black!0} {\scriptsize \href{https://osf.io/gvdse/}{CSV, \textcolor{black}{MP4}}} \tabularnewline
    
    \cmidrule(lr){2-19}

    {\scriptsize \cite{bauersfeld2021} \tiny{{NeuroBEM}}} & \cellcolor{black!0} {\scriptsize $\sim$75'} & \cellcolor{green!10} {\scriptsize Real} & \cellcolor{black!0} {\scriptsize n/a} & \cellcolor{red!10} {\scriptsize \textcolor{black}{\xmark}} & \cellcolor{red!10} {\scriptsize \xmark} & \cellcolor{red!10} {\scriptsize \xmark} & \cellcolor{green!10} {\scriptsize 18m/s} & \cellcolor{red!10} {\scriptsize \xmark} & \cellcolor{red!10} {\scriptsize \xmark} & \cellcolor{black!0} {\scriptsize \xmark} & \cellcolor{red!10} {\scriptsize \xmark} & \cellcolor{red!10} {\scriptsize \xmark} & \cellcolor{green!10} {\scriptsize @1000Hz} & \cellcolor{green!10} {\scriptsize @400Hz} & \cellcolor{red!10} {\scriptsize \xmark} & \cellcolor{green!10} {\scriptsize @1000Hz} & \cellcolor{green!10} {\scriptsize @400Hz} & \cellcolor{black!0} {\scriptsize \href{https://download.ifi.uzh.ch/rpg/NeuroBEM/}{CSV}} \tabularnewline
    
    \bottomrule
\end{tabular}
}
 \newline \hfill \newline
 {\footnotesize
 \hypertarget{hyref:01}{$^\dag$}Bounding boxes, top-bottom left-rigth corners.
 \hspace{1em}
  \hypertarget{hyref:99}{$^{\diamond}$}Piloted.
 \hspace{1em}
 \hypertarget{hyref:09}{$^{\P}$}Autonomous.
 \hspace{1em}
 \hypertarget{hyref:02}{$^{*}$}Stereo.
 \hspace{1em}
  \hypertarget{hyref:07}{$^{\S}$}Leica laser tracker.
 \hspace{1em}
  \hypertarget{hyref:06}{$^{\ddag}$}9300 frames.\\
 \hspace{1em}
 \hypertarget{hyref:03}{$^{||}$}Internal corners.
\hypertarget{hyref:04}{$^{\dag\dag}$}Synthetic camera images.
 \hspace{1em}
 \hypertarget{hyref:05}{$^{\P\P}$}Area of interest of the gaze.
  \hspace{1em}
 \hypertarget{hyref:08}{$^{**}$}Simulated.
 }
 \vspace{-2em}
     \label{tab:related}
\end{table*}

Today, several conferences and journals, including NeurIPS, the IEEE Robotics and Automation Letter, and the International Journal of Robotics Research explicitly solicit data and benchmark papers as a way to increase the number and visibility of peer-reviewed datasets~\cite{newman2009data}. 

Robotics datasets and benchmarks are as important and beneficial to the community as they are challenging to create---because of the idiosyncrasies of robotic hardware and real-world systems.
Notable robotics datasets have focused on vision problems, e.g., the KITTI Vision Benchmark Suite~\cite{Geiger2013IJRRkitti}, including data from stereo cameras, GPS, and laser scanners for tasks such as object detection, tracking, and visual-inertial odometry (VIO)~\cite{upenn2018}, or the use of special, novel sensors and instruments~\cite{wenda2022}.
Early drone racing datasets~\cite{alphapilotchallenge} also focused on scene understanding and gate pose estimation problems, while more recent datasets have put a greater emphasis on the coupling with on-board inertial data, ground truth information, and controls~\cite{hanover2023autonomous,loquercioscirob}.

Our dataset builds upon these and aims to be a one-stop resource for researchers to simultaneously pursue multiple lines of work, including semantic scene understanding, VIO, mapping and planning, and data-based {system identification} for fast and aggressive multi-rotor flight.
As for a benchmark to be successful, it must be effectively and easily repeatable, in Section~\ref{sec:system}, we release the complete design specifications of the drone used to collect the dataset.

The main contributions of our work are as follows. 
\begin{itemize}
    \item The public release of a dataset for drone racing (inclusive of open-source code for visualization and post-processing) that is characterized by:
    \begin{itemize}
        \item fast ($>$20m/s), aggressive flight, both autonomous and human-piloted, on multiple trajectories {
        (including a complex 3D racing track);
        }
        \item high-resolution, high-frequency ($\sim$10$^2$Hz) collection of visual, inertial, and motion capture data;
        \item versatility---our dataset includes drone racing gates fully labeled to the level of individual corners~\cite{scaramuzza-auro} (for VIO, self-localization, scene understanding, etc.), information about commands, control inputs, and battery voltages (for {estimation problems, etc.), as well as lighting and sensor settings metadata.}
    \end{itemize}
    \item The open design (with commercial off-the-shelf (COTS) components) of the racing drone used to collect the data. For direct comparisons, the same design allows, without modifications, both autonomous and piloted flight.
\end{itemize}
 \section{Related Work}
\label{sec:related}

In Table~\ref{tab:related}, we summarize the---all very recent---datasets for vision-based flight, drone racing, and aggressive quadrotor control related to our own.
Earlier datasets for multi-rotor VIO and simultaneous localization and mapping (SLAM) included the 2016 
EuRoC~\cite{burri2016} and the 2017 Zurich Urban~\cite{zurichmav2017} micro aerial vehicle (MAV) datasets.
However, these were characterized by comparatively lower speeds and frequencies of images and collected data than those needed for drone racing.

The last five years have seen a renewed interest in aggressive flight~\cite{hanover2023autonomous}.
In~\cite{upenn2018}, a new dataset was introduced to validate stereo VIO methods for fast autonomous flight, although without the inclusion of racing gates.
The Blackbird dataset~\cite{antonini2020} was proposed as an aggressive indoor flight dataset for agile perception.
While inertial data were collected in the real world, its high-resolution images were generated in simulation. 
In 2019, Lockheed Martin released an image dataset for Test\#2 of its AlphaPilot challenge's virtual qualifiers~\cite{alphapilotchallenge}, that did not include drone state information (while Test\#3 consisted of control in simulation).

The work closest to ours is the dataset presented in~\cite{delmerico2019}. It also observes that previous benchmarks for drone VIO had focused on too slow trajectories.
Our dataset differs from~\cite{delmerico2019} in that it provides 
higher-frequency RGB mono-camera images (the information used by human pilots), {it includes piloted and autonomous flights on identical tracks}, and it is fully annotated with high-frequency motion capture data as well as the gates' corner labels.

Other recent, specialized datasets for drone racing and aggressive multi-rotor flight include~\cite{pfeiffer2021,bauersfeld2021}.
Another open-source, open-hardware racing drone was proposed in~\cite{foehnscirob}. 
{In comparison, our design has a more powerful companion computer and it can be seamlessly used by human pilots.}

Compared to the existing literature (Table~\ref{tab:related})~\cite{hanover2023autonomous}, our dataset
\emph{(i)} includes high-resolution, high-frequency images captured in different light settings and \emph{(ii)} it is fully annotated, down to the gates' individual corners.
It can support research in all aspects of autonomous drone racing from VIO, to gate pose estimation~\cite{WinnerAlphapilot, Kaufmann2018BeautyAT} and data-driven control.
 \section{An Open-design Quadrotor for Autonomous Drone Racing and Data Collection}
\label{sec:system}

To collect this dataset, we designed a new custom quadrotor (Figure~\ref{fig:drone}).
Our design is based on a 5'' carbon-fiber frame with a propeller-to-propeller diagonal of 215mm. 
The fully-assembled drone, including the battery, weighs $\sim$870g{, has a thrust-to-weight ratio of 7.5 (load cell),}
and can reach a maximum speed {(measured outdoors)} of 179km/h---allowing for the aggressive maneuvers required in drone racing. {The top linear acceleration and angular velocity in the dataset are 69.85m/s$^2$ ($>$7$g$'s) and 20.01rad/s}.
Importantly, our design can be used, without modifications, as both an autonomous and a human-piloted FPV racing drone. 
We do this to create a true test bench to benchmark drone racing autonomy against human performance.
Our dataset includes both autonomous and piloted flights.
Its components are listed in Table~\ref{tab:platform}.

\subsection{Design Overview}

\begin{table}[h]
    \centering
    \caption{
    Off-the-shelf components needed to re-create the open-design racing drone used to collect the dataset
    }
    \begin{tabular}{
>{\raggedleft}p{1.6cm} >{\centering}p{1.5cm}
>{\centering}p{4.1cm}
}
\toprule

\textbf{Component} & \textbf{Producer} & \textbf{Model} \tabularnewline \cmidrule(lr){1-3}

ESC & T-MOTOR & F55A PROII 6S 4IN1 
\tabularnewline  \cmidrule(lr){2-3}

FCU & Holybro & Kakute H7 v1.3 
\tabularnewline  \cmidrule(lr){2-3}

\multirow{2}{*}{RC Receiver} & Team BlackSheep & CROSSFIRE NANO RX (SE) LONG RANGE
\tabularnewline  \cmidrule(lr){2-3}

\multirow{2}{*}{Battery} & \multirow{2}{*}{Tattu} & R-Line v5.0 1400mAh 22.2V 150C 6S1P LiPo
\tabularnewline  \cmidrule(lr){1-3}

Computer & \multirow{1}{*}{NVIDIA} & \multirow{1}{*}{Orin NX 16GB Module}
\tabularnewline  \cmidrule(lr){2-3}

Carrier board & Seeed Studio & A203 (Version 2) 
\tabularnewline \cmidrule(lr){2-3}

BEC & Matek & BEC12S-PRO 
\tabularnewline  \cmidrule(lr){2-3}

Camera & Arducam & \textcolor{black}{B0179} 8MP IMX219 \tabularnewline  \cmidrule(lr){1-3}

FPV Camera & Foxeer & T-Rex Mini 1500TVL
\tabularnewline \cmidrule(lr){1-3}

\multirow{1}{*}{Frame} & \multirow{1}{*}{Pyrodrone} & Hyperlite Floss 3.0 Race Frame 5"
\tabularnewline  \cmidrule(lr){2-3}

Motors & T-MOTOR & F60 PRO V 2020KV 
\tabularnewline \cmidrule(lr){2-3}

Propellers & T-MOTOR & T5147 
\tabularnewline

\bottomrule
\end{tabular}

     \label{tab:platform}
\end{table}

The quadrotor has three main sub-systems: \emph{(i)} the quadrotor electronics, \emph{(ii)} the autonomous module, and \emph{(iii)} the First-Person-View (FPV) system. 
These sub-systems are combined by means of the frame and fasteners.
The assembled system comprises two cameras. 
One digital, connected to the autonomous module, and one analog, used by the human pilot in the FPV system. 
The two cameras share the same mount, and the FPV camera is placed above the digital one (Figure~\ref{fig:drone}).

\subsubsection{Quadrotor electronics}
\label{sec:quad_electronics}
These are \emph{(i)} the electronic speed controller (ESC), \emph{(ii)} the Kakute H7 v1 flight controller unit (FCU), \emph{(iii)} the radio controller (RC) receiver, and \emph{(iv)} the battery.
They are mounted underneath the frame. 
These components are protected by the aluminum standoffs connecting the frame and a 3D-printed custom battery cage. 
The FCU hosts an STM32H7 microcontroller and it is capable of running multiple firmware, including Betaflight, Ardupilot, and PX4.

\subsubsection{Autonomous module}
\label{sec:auto_module}
It comprises \emph{(i)} an NVIDIA Orin NX (hosted on the A203v2 carrier board with SSD and wireless card), \emph{(ii)} the battery eliminator circuit (BEC) powering it, and \emph{(iii)} an Arducam RGB camera.
These components are placed above the frame and are secured by two 3D-printed plates, connected by aluminum standoffs. 
The top plate provides the mount for the cameras. A MIPI CSI-2 ribbon cable connects the companion board with the Arducam. The FC is connected via a serial port, using a shielded cable. This connection is used to both control the drone and read FC's sensors.

\subsubsection{FPV system}
\label{sec:fpv_system}
Independent from the autonomous module and used for human piloting instead, it comprises an FPV analog camera, a video transmitter, and its antennas, all placed above the frame.

The CAD models of all the 3D printed parts and the bill of materials of all other COTS components are available online\footnote{
\href{https://github.com/tii-racing/drone-racing-dataset/tree/main/quadrotor}{\texttt{github.com/tii-racing/drone-racing-dataset/tree/} \texttt{main/quadrotor}}
}, with a video tutorial on how to re-create our drone.

\begin{figure}
    \centering
    \includegraphics[]{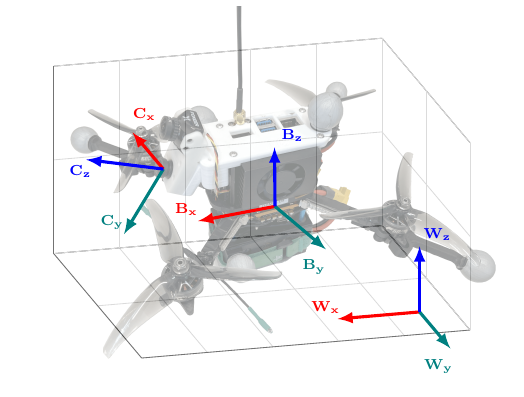}
    \caption{
    The drone platform used to record the dataset, the body frame $B$ has its origin at the FCU's IMU location, the camera frame $C$ is located where the bottom lens is (the top lens being the one of the FPV system).
    }
    \label{fig:drone}
\end{figure}

\subsection{Sensors}

Our quadrotor is equipped with multiple sensors for autonomous and piloted aggressive flight:

\subsubsection{InvenSense MPU6000 IMU}
Part of the quadrotor electronics (\ref{sec:quad_electronics}), it is embedded into the FC. 
This module has two functions: delivering precise, real-time tri-axis angular rate sensor (gyroscope) data, as well as accurate tri-axis accelerometer data.
The raw IMU data are read by the companion computer in a demand/response exchange fashion, using the Multiwii Serial Protocol (MSP)~\cite{mspprotocol}.

\subsubsection{Arducam {B0179} IMX219 8MP RGB Bayer camera}
Part of the autonomous module (\ref{sec:auto_module}), it captures 640$\times$480 pixel frames at 120Hz with a diagonal field-of-view (FOV) of 175\textdegree ~{and a horizontal field-of-view (HFOV) of 155\textdegree}. {Its readout speed is 3.22$\times $10$^{-5}$s, computed as the line length divided by the pixel rate before re-scaling, i.e., 3560px/(1280$\times$720$\times$120Hz).}
It is one of the most used lightweight embedded cameras on the market and it is fully supported by NVIDIA with dedicated MIPI CSI-2 drivers.
The image YUV frames are captured in NV12 format. In the NV12 format, each pixel in the luminance (Y) component is represented by a single byte. The chrominance (UV) components are interleaved and share memory locations. The image is then converted to BGR, a common color format suitable for processing and analysis. Eventually, the image is saved as JPEG. 
This pipeline is accomplished by the NVIDIA GStreamer plugin on the NVIDIA companion computer.

\subsubsection{Foxeer T-Rex Mini 1500TVL}
It is the low latency (6ms) camera in the FPV system (\ref{sec:fpv_system}).
For the sake of the data collection in this letter, this was used by a human pilot in conjunction with a pair of 1280x960 OLED Fat Shark HDO2 googles.

\subsection{Software}

\subsubsection{Quadrotor electronics (\ref{sec:quad_electronics}) software}
The FC firmware we used is Betaflight 4.3.1~\cite{betaflight}, whose proportional-integral-derivative controller (PID) was tuned by a human pilot. 
The companion computer uses MSP to both send commands to the FC and read its sensors.
Accordingly, we activate Betaflight's \texttt{MSP\_OVERRIDE} feature to bypass the RC controller's commands. An \texttt{MSP\_OVERRIDE} channels mask limits to override the motor commands: for safety reasons, a human can always disarm the drone with an RC controller.

\subsubsection{Autonomous module (\ref{sec:auto_module}) software}
On the Orin NX module, we installed NVIDIA JetPack 5.1.1. The JetPack includes Jetson Linux 35.3.1 Board Support Package (BSP) with Linux Real-Time Kernel 5.10, an Ubuntu 20.04-based root file system with CUDA 11.4 support. 
We use the Humble (current LTS) distribution of the Robot Operating System 2 (ROS2) as the middleware for communication between the perception, planning, and control modules. 

\begin{figure}
    \centering
    \includegraphics[trim={0cm 1.5cm 0cm 0cm},clip]{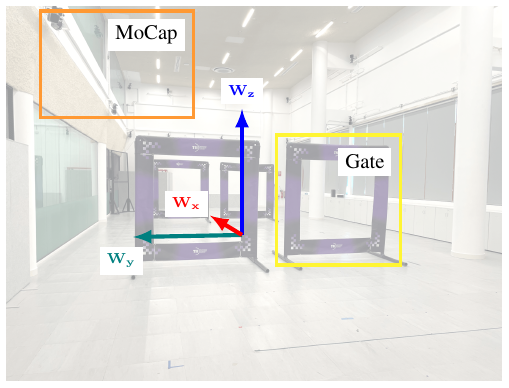}
    \caption{
    The 25$\times$9.7$\times$7 meters indoor arena, instrumented with 32 Qualisys MoCap cameras and equipped with four 5$\times$5 feet racing gates used to record the dataset.
    }
    \label{fig:arena}
\end{figure} \section{Data Collection Protocol}
\label{sec:experiments}

\subsection{Flight Arena, Racing Gates, and Motion Capture System}
Our dataset was recorded in a 25 (L) by 9.7 (W) by 7 (H) meters indoor flying arena. 
The arena is equipped with a 32-camera Arqus A12 Qualisys Motion Capture (MoCap) system, tracking 6DoF poses of defined rigid bodies with millimeter accuracy at 275Hz. 
The drone design from Section~\ref{sec:system} is equipped with six 25mm markers defining a single rigid body.
The markers are mounted on the top plate, battery cage, and 48.2mm arm extensions, preventing the markers from being occluded by the propellers (Figure~\ref{fig:drone}).
The origin of the drone's rigid body was placed at the location of the FCU's IMU, as shown in Figure~\ref{fig:drone}.
The IMU was also calibrated using the RC before each take-off~\cite{betaflight}.

{
We use a minimum of 4 and up to 7 gates to create 2D and 3D racing tracks.
The 7-gate track is inspired by the one in~\cite{scaramuzza-nature} but shrunken to meet our arena size constraints (Figure~\ref{fig:arena}).
This track features challenging maneuvers like the Split-S and sharp turns, and it evolves over the $z$-axis for 5 meters.
} 
The gates are made of PVC pipes covered by printed fabric banners.
They measure 7 by 7 feet (213.36 cm) and have an internal opening of 5 by 5 feet (152.4 cm), similar to those used in major drone racing leagues~\cite{gatesize}.
Each racing gate's rigid body is defined by four markers placed in its inner corners.

\subsection{Flight Program}
\begin{figure*}
    \centering
\scalebox{0.975}{
    \includegraphics[]{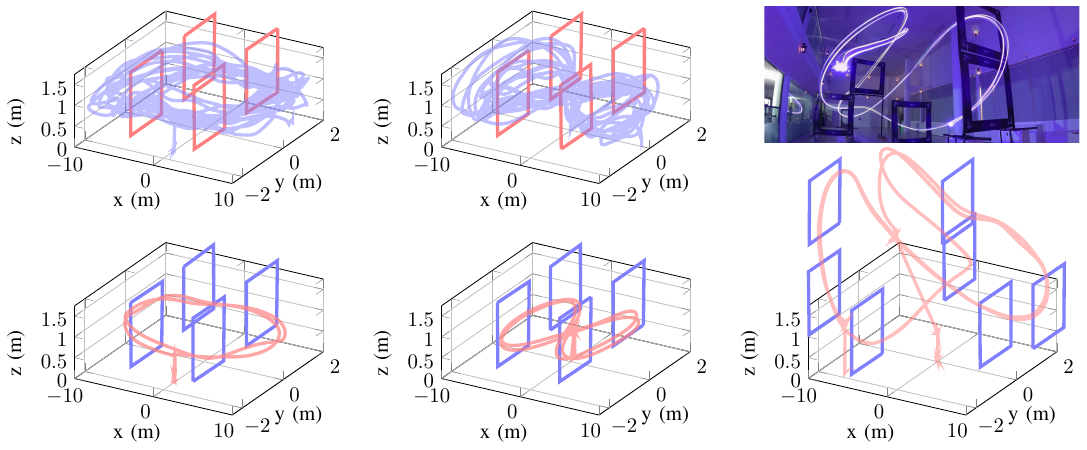}
    }
    \caption{
    {
    Examples of recorded trajectories: 
    piloted ellipse \texttt{\small flight-01p-ellipse} (top-left),
    piloted lemniscate \texttt{\small flight-07p-lemniscate} (top-center),
    autonomous ellipse \texttt{\small flight-01a-ellipse} (bottom-left),
    autonomous lemniscate \texttt{\small flight-07a-lemniscate} (bottom-center),
    and autonomous race track \texttt{\small flight-13a-trackRATM} (bottom-right).
    }
    \vspace{-.6cm}
    }
    \label{fig:trajectory3d}
\end{figure*}

Our dataset contains a total of {30} flights (Table~\ref{tab:flight_summary}): 12 \emph{human-piloted} and {18} \emph{autonomous} ones. 
In either case, two shapes---\emph{ellipse} and \emph{lemniscate} (Figure~\ref{fig:trajectory3d})---have been executed 6 times. 
{The 6 additional autonomous flights are collected on a 3D race track.}
The 6 repetitions correspond to the combination of 3 different illumination conditions and 2 camera settings.

\begin{table}[t]
    \centering
    \caption{
    Summary of the flights recorded in the dataset
    }
    \begin{tabular}{ 
c
c
c
c
c
} 

\toprule

\textbf{Control} & 
\textbf{Shape} &
\textbf{Top Speed} &
\textbf{Time} &
\textbf{Distance}
\tabularnewline 

\cmidrule(lr){1-5}

\multirow{3}{*}{Autonomous} & 
Ellipse\hyperlink{hyref:21}{$^{\dag}$}    &
\textbf{21.83 m/s} &
149.08 s &
455.27 m
\tabularnewline

 &
Lemniscate\hyperlink{hyref:21}{$^{\dag}$} &
10.22 m/s &
155.08 s &
359.63 m
\tabularnewline 

 &
{Race Track\hyperlink{hyref:22}{$^{\ddag}$}} &
{21.39 m/s} &
{278.05 s} &
{1161.51 m}
\tabularnewline 

\cmidrule(lr){1-5}

\multirow{2}{*}{Piloted} &
Ellipse\hyperlink{hyref:23}{$^{\S}$}       &
9.50 m/s &
575.38 s &
2586.3 m
\tabularnewline

 &
Lemniscate\hyperlink{hyref:23}{$^{\S}$}    &
8.93 m/s &
\textbf{593.63 s} &
\textbf{2593.47 m}
\tabularnewline 

\bottomrule

\end{tabular}
 \newline \hfill \newline
 {\footnotesize
 \hypertarget{hyref:21}{$^\dag$}Flown twice in 6 flights (3 brightness $\times$ 2 camera settings).
   {\hypertarget{hyref:22}{$^{\ddag}$}Flown three times in 6 flights (3 brightness $\times$ 2 camera settings).}
  \hypertarget{hyref:23}{$^\S$}Flown as many times as possible in 6 flights (3 brightness $\times$ 2 camera settings).
 }
     \label{tab:flight_summary}
\end{table}

\subsubsection{Brightness levels} 
We created three levels of brightness for data collection. The \emph{high} brightness level is achieved with both natural and artificial light {(max./avg./min. measured illuminance in the arena of 2480 lx, 1500 lx, and 254 lx)}; the \emph{medium} level has controlled light, and it is obtained by turning on all the artificial lights in the arena but keeping natural light out using the blinds {(955 lx/672.3 lx/254 lx)}. The \emph{low} brightness condition is obtained by turning off most of the lights in the arena and keeping the blinds down, {(216 lx/72 lx/34.2 lx)}.

\subsubsection{Camera settings} 
We used two different {nvarguscamerasrc} settings, \emph{auto} exposure time and gains{,} and {\emph{fixed} exposure time (2.5ms), analog gain (2), and digital gain (1)}. In the \emph{auto} setting, the image is brighter but suffers from higher motion blur. In \emph{fixed}, the image is darker, with limited motion blur.

\subsection{Human-piloted and Autonomous Control}
In autonomous mode, each flight contains exactly two {(ellipse, lemniscate)  or three (race track) laps, lasting $\sim$25 and $\sim$45s, respectively.}
The human-piloted flights last between 84 and 108{s}, during which the pilot tries to achieve the maximum number of laps possible on a single battery charge.
{We did not clip the human-piloted flights as they exhibit higher variability and cover a larger state space.}

In the autonomous flights, we use a Proportional Derivative (PD) controller based on~\cite{8118153} {for the 2D tracks and a Model Predictive Controller (MPC) based on~\cite{8593739} for the 3D track}.
The motion capture system feeds the current quadrotor pose into the controller {at 275Hz} through WiFi.
{Control commands are sent to the FCU at the same update frequency of 275Hz by the PD controller, and 160Hz by the MPC. A Python implementation of the PD controller and its tuned gains are available on GitHub}\footnote{
\href{https://github.com/tii-racing/drone-racing-dataset/blob/main/scripts/reference_controller.py}{\texttt{github.com/tii-racing/drone-racing-dataset/blob/} \texttt{main/scripts/reference\_controller.py}}
}.

The human pilot used a camera angle (for both their FPV camera and the recorded Arducam) of 30\textdegree.
The autonomous runs were recorded with a camera angle of 40\textdegree\ for the lemniscate and 50\textdegree\ for the ellipse { and race track, empirically chosen to maximize the gates' visibility}. 
FPV pilots choose the camera angles according to the speed they plan to reach\footnote{
\href{https://www.getfpv.com/learn/fpv-essentials/fpv-camera-angle-full-throttle-flight/}{{\texttt{www.getfpv.com/learn/fpv-essentials/fpv-camera-} \texttt{angle-full-throttle-flight/}}}
}. We used the same principle to choose the camera angle for the autonomous mode, making the gates visible at high speed.

\subsection{Image Labeling}
Often being the only well-known object in a racing environment, gates are key to relative localization and next-waypoint detection.
In this dataset, we provide image labels in the form of bounding boxes and keypoints associated with the inner corners of all visible gates.
Coupled with the drone's inertial data and the ground truth from the motion capture system, our dataset allows to reproduce and benchmark the state-of-the-art in gate pose estimation~\cite{Pham2022DeepLF} as well as to develop new methods.

Labeling was first performed automatically, using a top-down keypoints detector~\cite{chen2019mmdetection,mmpose2020} trained on a synthetic dataset and fine-tuned on 5'000 manually labeled images. 
All the images were then labeled through an iterative process of manual review and re-training.
Finally, all labels were eventually manually reviewed.
Bounding boxes and corner positions are provided, based on the auto-labeling results and a human estimate, even for partially occluded gates.
However, no distinction between occluded and fully visible keypoints is made. Thus, the only visibility values we provide are 0 (outside the image boundaries) and 2 (inside the image boundaries) as per COCO format definition~\cite{cocoformat}.
As a convention, gates are not labeled when there are no visible corners.

\subsection{Time Synchronization}
We record three separate data streams during each flight: \emph{(i)} a \texttt{\small rosbag} containing the FCU readings and the autonomous control setpoints, \emph{(ii)} the on-board camera images, and \emph{(iii)} the Qualisys motion capture measurements.

For the FCU data, we use custom ROS2 messages and the Real-Time Kernel to limit the jitter in the sensors' readings.
Furthermore, the GStreamer pipeline allows us to save the images and timestamp them using the frame acquisition time.

In the end, two different clocks are involved: the real-time clock of the drone's companion computer, and the clock of the Qualisys workstation. Both were synced with a Network Time Protocol (NTP) server placed inside the facility, before each flight.
The clock offsets w.r.t. the NTP server of the two machines were recorded before and after each flight to compute 
the jitter that occurred during the flight.
The drone achieved a microsecond clock accuracy with Chrony, while the Qualisys workstation recorded a millisecond accuracy.
The total jitter from start to end of a trajectory never exceeded 3ms.

\subsection{Data Post-processing}
The motion capture data were converted to CSV, and the clock offset with the onboard computer was removed. The ROS2 bags were also dumped into CSVs. All the data were then trimmed to remove pre-take-off and post-landing records.
We used an open-source script for data alignment (Snipped~\ref{alg:interpolation}) to produce user-friendly comprehensive CSVs. 
The alignment is achieved through linear interpolation for all the fields with the exception of the rotation matrices, for which spherical linear interpolation~\cite{slerp10.1145/325334.325242} is used instead.

 \section{Data Format}
\label{sec:data}

For each flight, we recorded data from four different sources: \emph{(i)} the motion capture system's ground truth (drone and gates poses), \emph{(ii)} the Arducam IMX219 Bayer camera (images), \emph{(iii)} the flight controller (IMU, battery, motors, RC), and \emph{(iv)} companion computer (autonomous control reference and control inputs). 

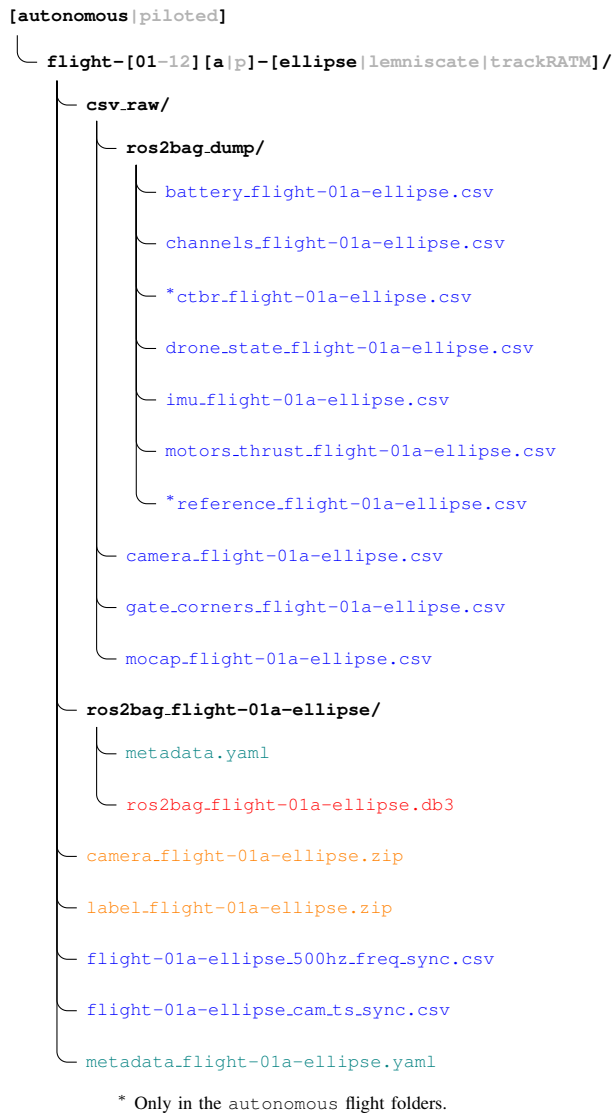
\begin{figure}[hbtp!]
\begin{center}
\begin{forest}
  for tree={
    font=\scriptsize\ttfamily,
rounded corners=4pt,
    grow'=0,
    child anchor=west,
    parent anchor=south,
    anchor=west,
    calign=first,
edge={black,rounded corners,line width=.4pt},
    edge path={
      \noexpand\path [draw, \forestoption{edge}]
      (!u.south west) +(7.5pt,0) |- (.child anchor)\forestoption{edge label};
    },
    before typesetting nodes={
      if n=1
        {insert before={[,phantom]}}
        {}
    },
    fit=band,
    before computing xy={l=15pt},
  }
[\url{https://github.com/tii-racing/drone-racing-dataset}\textbf{/data/}
    [{\bfseries {[}autonomous\textcolor{black!30}{|piloted}{]}}
        [{\bfseries flight-{[}01\textcolor{black!30}{-12}{]}{[}a\textcolor{black!30}{|p}{]}-{[}ellipse\textcolor{black!30}{|lemniscate|trackRATM}{]}/}
            [{\bfseries csv\_raw/}
                [{\bfseries ros2bag\_dump/}
                    [\textcolor{blue!100}{battery\_flight-01a-ellipse.csv}]
                    [\textcolor{blue!100}{channels\_flight-01a-ellipse.csv}]
                    [\textcolor{blue!100}{\hyperlink{hyref:31}{$^*$}ctbr\_flight-01a-ellipse.csv}]
                    [\textcolor{blue!100}{drone\_state\_flight-01a-ellipse.csv}]
                    [\textcolor{blue!100}{imu\_flight-01a-ellipse.csv}]
                    [\textcolor{blue!100}{motors\_thrust\_flight-01a-ellipse.csv}]
                    [\textcolor{blue!100}{\hyperlink{hyref:31}{$^*$}reference\_flight-01a-ellipse.csv}]
                ]
                [\textcolor{blue!100}{camera\_flight-01a-ellipse.csv}]
                [\textcolor{blue!100}{gate\_corners\_flight-01a-ellipse.csv}]
                [\textcolor{blue!100}{mocap\_flight-01a-ellipse.csv}]
            ]
            [{\bfseries ros2bag\_flight-01a-ellipse/}
                [\textcolor{teal!100}{metadata.yaml}]
                [\textcolor{red!100}{ros2bag\_flight-01a-ellipse.db3}]
            ]
            [\textcolor{orange!100}{camera\_flight-01a-ellipse.zip}]
            [\textcolor{orange!100}{label\_flight-01a-ellipse.zip}]
            [\textcolor{blue!100}{flight-01a-ellipse\_500hz\_freq\_sync.csv}]
            [\textcolor{blue!100}{flight-01a-ellipse\_cam\_ts\_sync.csv}]
            [\textcolor{teal!100}{metadata\_flight-01a-ellipse.yaml}]
        ]
    ]
]
\end{forest}

{\scriptsize \hypertarget{hyref:31}{$^*$} Only in the \texttt{autonomous} flight folders.}
\end{center}
    \caption{
    Folder and file structure of the dataset. 
    }
    \label{fig:tree}
\end{figure}

All data collected is timestamped (Unix epoch time) with a microsecond resolution.
In every flight folder, a YAML metadata file summarizes the camera and light setting, along with the type of track. 
The total time of flight and meters traveled are also included.
The folder and file structure of the dataset are shown in Figure~\ref{fig:tree}

\subsection{Camera-aligned and Uniform-sampling CSVs}
\label{sec:csvs}

For each flight, we pre-compiled two easy-to-use, comprehensive CSV files that include all the data detailed in the subsequent sections, aligned via interpolation. 
In each \texttt{\small [FLIGHT]\_cam\_ts\_sync.csv} file, we use the timestamps from the camera frames, and all other data points are linearly interpolated to align with these timestamps. 
Conversely, in each \texttt{\small [FLIGHT]\_500hz\_freq\_sync.csv} file, timestamps are sampled at a uniform 500Hz frequency between the first and last camera timestamps. All numerical data are again interpolated and aligned with timestamps, and each row references the file name of the camera frame with the closest timestamp.
All the columns in each of these CSVs are presented in Table~\ref{tab:maincsv}.

\subsection{Drone and Gates' Poses from Motion Capture}

For each flight, files \texttt{\small gate\_corners\_[FLIGHT].csv} contain the timestamped x, y, and z coordinates of all the gates' markers, in meters. Files \texttt{\small mocap\_[FLIGHT].csv} contain the poses of all the rigid bodies, i.e., the drone and the gates. Each pose comprises x, y, z, roll, pitch, yaw, measurement residual, and the orientation described by a 3$\times$3 column-major order matrix. Poses are in meters and radiants.

\subsection{Camera Frames}

The images from the Arducam IMX219 Bayer camera are recorded at 120 FPS with a resolution of 640$\times$480px and saved as JPEG files. They are provided as a ZIP file along with a \texttt{\small camera\_[FLIGHT].csv} which contains the timestamps of the acquisition of the frame and the name of the corresponding JPEG file.

\subsection{Image Labels}

For each image, the gates' bounding boxes and internal corner labels are given as a TXT file with the same name as the JPEG file.
Each line in a TXT file represents a single gate in the form:
\[
0\,\, c_x\,\, c_y\,\, w\,\, h\,\, tl_x\,\, tl_y\,\, tl_v\,\, tr_x\,\, tr_y\,\, tr_v\,\, br_x\,\, br_y\,\, br_v\,\, bl_x\,\, bl_y\,\, bl_v
\]
where $0$ is the class label for a gate (the only class in our dataset); 
$c_x$, $c_y$, $w$, $h$ $\in [0, 1]$ are its bounding box center's coordinates, width, and height, respectively; and $tl_x$, $tl_y$ $\in [0,1]$, $tl_v$ $\in [0;2]$ are the coordinates and visibility (0 outside the image boundaries; 2 inside the image boundaries) of the top-left internal corner. Similarly for $tr$, $bl$, $br$, the top-right, bottom-left, and bottom-right corners.
All values are in pixel coordinates normalized with respect to image size.
The keypoints label format follows the COCO definition~\cite{cocoformat}. The labels are provided as a ZIP file in \texttt{\small label\_[FLIGHT].zip}.

\subsection{On-board Data from the Quadrotor Electronics}

From the FCU in Subsubsection~\ref{sec:quad_electronics}, we record the following measurements:
battery's voltage (Volts), at 50Hz (\texttt{\small battery\_[FLIGHT].csv});
IMU, i.e., accelerometer ($m/s^2$) and gyroscope ($rad/s$) x, y, and z axes in the East-North-Up (ENU) board frame, at 500Hz (\texttt{\small imu\_[FLIGHT].csv});
single motor thrust feedback, normalized between 0 and 1, for all four motors, at 100Hz (\texttt{\small motors\_thrust\_[FLIGHT].csv});
RC's channels, i.e. roll, pitch, thrust, yaw, aux1, aux2, aux3, and aux4 values of the sticks between 1000 and 2000, at 100Hz (\texttt{\small channels\_[FLIGHT].csv}, only for human-piloted flights).

\subsection{On-board Data from the Autonomous Module}

From the NVIDIA Orin in Subsubsection~\ref{sec:auto_module}, we record the following measurements:
the drone state, i.e., position ($m$), orientation (quaternion), velocity ($m/s$), and angular velocity ($rad/s$), at 275Hz, sent as ground truth from the MoCap system to the drone, with the time delay of the communication channel (\texttt{\small drone\_state\_[FLIGHT].csv}).

Furthermore, only for the autonomous flights, we also record:
the controller's reference, i.e., position ($m$), orientation (quaternion), linear ($m/s$) and angular velocity ($rad/s$), acceleration ($m/s^2$), jerk ($m/s^3$), heading ($rad$), and heading rate ($rad/s$), computed at 100Hz {for the ellipse and lemniscate, and 500Hz for the race track} (\texttt{\small reference\_[FLIGHT].csv});
the collective thrust and body rates (CTBR) computed by the autonomous controller, i.e., the normalized thrust ($N/kg$ i.e. $m/s^2$), roll, pitch, and yaw rates ($rad/s$), at 275Hz {for the PD controller, and 160Hz for the MPC}, (\texttt{\small ctbr\_[FLIGHT].csv});
and the RC's channels sent to the FCU calculated from the CTBR commands, i.e., roll, pitch, thrust, yaw, aux1, aux2, aux3, and aux4 values of the sticks, at 275Hz {and 160Hz for PD and MPC respectively}, (\texttt{\small channels\_[FLIGHT].csv}).  

\begin{table}[h]
\centering
\caption{
Data available in the precompiled \texttt{\small cam\_ts\_sync.csv} and \texttt{\small 500Hz\_freq\_sync.csv} (Sec.~\ref{sec:csvs})
}
\scalebox{0.95}{
\begin{tabular}{
>{\raggedright}p{5cm}
>{\centering}p{.5cm}
>{\centering}p{2.2cm}
} 

\toprule

\textbf{Column Number and Quantity Name} &
\textbf{Unit} &
\textbf{Data Type}
\tabularnewline

\cmidrule(lr){1-3}

0. \texttt{\scriptsize elapsed\_time} &
$s$ &
float
\tabularnewline

1. \texttt{\scriptsize timestamp} &
$\mu s$ &
int
\tabularnewline

2. \texttt{\scriptsize img\_filename} &
n/a &
string
\tabularnewline

3. \texttt{\scriptsize accel\_[x\textcolor{black!30}{$|$y$|$z}]}                                            & $m/s^2$ & float\tabularnewline

6. \texttt{\scriptsize gyro\_[x\textcolor{black!30}{$|$y$|$z}] }                                            & $rad/s$ & float\tabularnewline

9. \texttt{\scriptsize thrust[0\textcolor{black!30}{-3}]     }                                            & $1$ & float $\in [0,1]$\tabularnewline

13. \texttt{\scriptsize channels\_[roll\textcolor{black!30}{$|$pitch$|$thrust$|$yaw}]}                       & $1$ & int $\in [1000,2000]$\tabularnewline

17. \texttt{\scriptsize aux[1\textcolor{black!30}{-4}]          }                                            & $1$ & int $\in [1000,2000]$\tabularnewline

21. \texttt{\scriptsize vbat                                    }                                            & $V$ & float\tabularnewline

22. \texttt{\scriptsize drone\_[x\textcolor{black!30}{$|$y$|$z}]}                                            & $m$ & float\tabularnewline

25. \texttt{\scriptsize drone\_[roll\textcolor{black!30}{$|$pitch$|$yaw}] }                                  & $rad$ & float\tabularnewline

28. \texttt{\scriptsize drone\_velocity\_linear\_[x\textcolor{black!30}{$|$y$|$z}]}                          & $m/s$ & float\tabularnewline

31. \texttt{\scriptsize drone\_velocity\_angular\_[x\textcolor{black!30}{$|$y$|$z}] }                        & $rad/s$ & float\tabularnewline

34. \texttt{\scriptsize drone\_residual                         }                                            & $m$ & float\tabularnewline

35. \texttt{\scriptsize drone\_rot[[0\textcolor{black!30}{-8}]] }                                            & $1$ & float\tabularnewline

44. \texttt{\scriptsize gate[1\textcolor{blue!100}{-7}]\_int\_[x\textcolor{black!30}{$|$y$|$z}]}             & $m$ & float\tabularnewline

56. \texttt{\scriptsize gate[1\textcolor{blue!100}{-7}]\_int\_[roll\textcolor{black!30}{$|$pitch$|$yaw}]}    & $rad$ & float\tabularnewline

68. \texttt{\scriptsize gate[1\textcolor{blue!100}{-7}]\_int\_residual }                                     & $m$ & float\tabularnewline

72. \texttt{\scriptsize gate[1\textcolor{blue!100}{-7}]\_int\_rot[[0\textcolor{black!30}{-8}]] }             & $1$ & float\tabularnewline

108. \texttt{\scriptsize gate[1\textcolor{blue!100}{-7}]\_marker[1\textcolor{black!30}{-4}]\_[x\textcolor{black!30}{$|$y$|$z}]}     & $m$ & float\tabularnewline

\bottomrule

\end{tabular}
}

 \label{tab:maincsv}
\end{table}

\begin{figure*}[hbt]
  \centering
  \scalebox{0.975}{
  \includegraphics[]{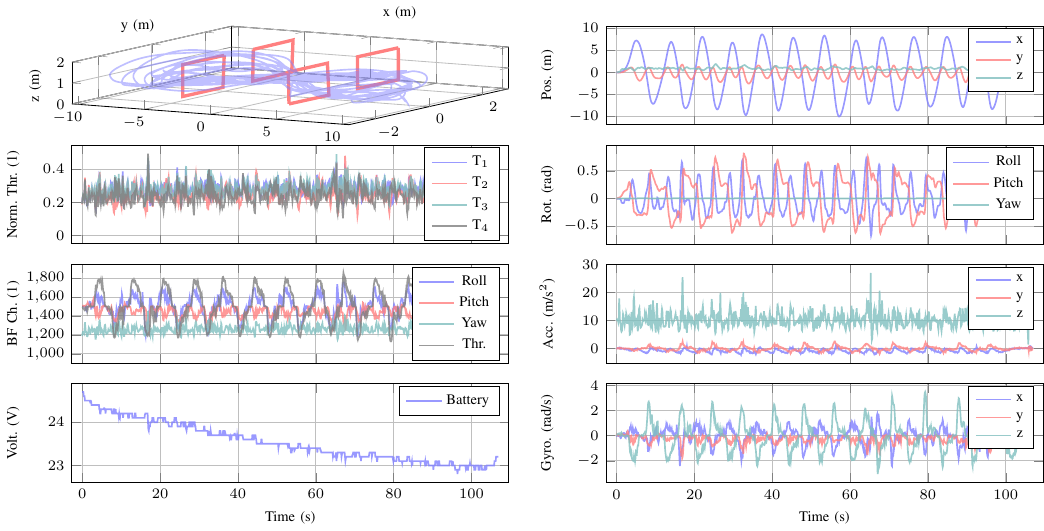}
  }
  \caption{
  Sample output of the utility script \texttt{\small data\_plotting.py} (Section~\ref{sec:scripts}), plotting 21 different data points for \texttt{\small flight-07p-lemniscate}, a piloted, 13-lap flight on the lemniscate trajectory.
  \vspace{-.6cm}
  }
  \label{fig:data_plotting}
\end{figure*}

\begin{figure}
    \hspace{-2.75mm}\centering
    \includegraphics[trim={0cm 0.5cm 0cm 1cm},clip,width=0.475\textwidth]{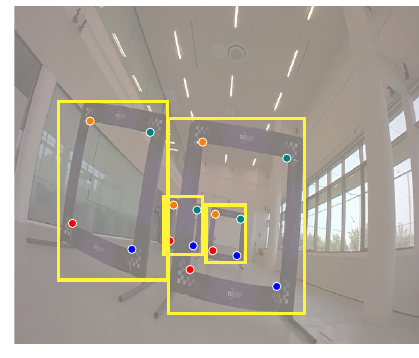}
    \caption{
    Sample output of script \texttt{\small label\_visualization.py} from Section~\ref{sec:scripts}, overlaying the bounding boxes and keypoints (top-left, top-right, bottom-right, and bottom-left corners) for all obstructed and unobstructed gates in one of the frames from \texttt{\small flight-02a-ellipse}.
    }
    \label{fig:label}
\end{figure}
 \section{Visualization and Post-processing Scripts}
\label{sec:scripts}

Along with the dataset, we provide an installable\footnote{
\href{https://github.com/tii-racing/drone-racing-dataset/blob/main/README.md}{\texttt{github.com/tii-racing/drone-racing-dataset/blob/} \texttt{main/README.md}}
}
open-source Python repository comprising of a set of processing scripts to visualize and manipulate the data.

\href{https://github.com/tii-racing/drone-racing-dataset/blob/main/scripts/data_interpolation.py}{\texttt{\small data\_interpolation.py}} can be used to re-sample and align the data at an arbitrary frequency using linear interpolation.
Its output is a CSV file containing the aligned, re-sampled data from all the sources in Section~\ref{sec:data}. First, one selects a flight by passing its id, e.g., \texttt{\small flight-01a-ellipse} as an argument.
Then, one can choose which synchronization option to use. In option \emph{1}, one must choose a new frequency for which timestamps are generated, using the first and last camera timestamp as a window. All the data are then interpolated to the generated timestamps. For the camera frames, the one with the closest timestamp is referred to by file name. In option \emph{2}, the timestamps of the camera CSV are used as a reference instead.

\begin{lstlisting}[firstnumber=1,
    numbers=none,
label = {alg:interpolation},]
$ python3 data_interpolation.py \
> --flight flight-01a-ellipse \
> --sync-option 1 --freq 200
Loading and pre-processing CSVs...
Using frequency 200 to interpolate all data
Syncing dataframes...
Sync complete.
Saving final CSV...
Final CSV saved.
\end{lstlisting}

\href{https://github.com/tii-racing/drone-racing-dataset/blob/main/scripts/data_plotting.py}{\texttt{\small data\_plotting.py}} is a script to simultaneously visualize multiple sensor data using customizable subplots. 
One can select which subplots to include by means of command-line arguments. 
The script facilitates quick insights into the drone's flight behavior and performance, such as its 3D trajectory and the gate positions, as well as pose, velocities, accelerations, battery voltage, RC channels, etc.; an example of the script output is shown in Figure~\ref{fig:data_plotting}.

\begin{lstlisting}[firstnumber=1,language=Bash,
    numbers=none,
label = {alg:plotting},]
$ python3 data_plotting.py \
> --csv-file flight-01a-ellipse_cam_ts_sync.csv \
> --subplots 3d
\end{lstlisting}

\href{https://github.com/tii-racing/drone-racing-dataset/blob/main/scripts/label_visualization.py}{\texttt{\small label\_visualization.py}} is the script to visualize the label annotations on the images. 
It allows one to skim the images of a flight, read the associated YOLO-style label file, and plot the gates' bounding boxes and internal corner keypoints. 
An individual frame example is shown in Figure~\ref{fig:label}.

\begin{lstlisting}[firstnumber=1,language=Bash,
    numbers=none,
label = {alg:label_viz},]
$ python3 label_visualization.py \
> --flight flight-01a-ellipse
\end{lstlisting}

\href{https://github.com/tii-racing/drone-racing-dataset/blob/main/scripts/create_std_bag.py}{\texttt{\small create\_std\_bag.py}} is a utility script that reads all the data from a user-specified flight and creates a new ROS2 bag with the same data (including images) but only using standard messages from the ROS2 \texttt{\small msgs} library. All the messages are timestamped in nanoseconds. {It may take several GBs.}

\begin{lstlisting}[firstnumber=1,language=Bash,
    numbers=none,
label = {alg:std_bag},]
python3 create_std_bag.py \
> --flight flight-01a-ellipse
\end{lstlisting}

 \section{Conclusions}
\label{sec:conclusions}

The development of autonomous racing drones requires simultaneously tackling challenging perception, state estimation, and control tasks---in real time---under limited computational resources.
With this dataset, we created a one-stop resource to develop and evaluate new algorithms for autonomous drone racing.
Our work is comprehensive of both aggressive autonomous and piloted flight; high-resolution, high-frequency visual, inertial, and motion capture data; commands and control inputs; multiple light settings; and corner-level labeling of drone racing gates.
Along with the data, we open-sourced the scripts used to parse and visualize them.
To further democratize autonomous drone racing research, we also released the parts list and instructions to recreate our flight platform, using commercial off-the-shelf components, hoping to see it re-created and used by researchers around the world.

\end{document}